%% file: main.tex
\documentclass[10pt,twocolumn,letterpaper]{article}
\usepackage[pagenumbers]{cvpr}
\usepackage[T1]{fontenc}
\usepackage[utf8]{inputenc}
\usepackage{microtype}
\usepackage{inconsolata}
\usepackage{graphicx}
\usepackage{amsmath,amssymb,booktabs,array}
\definecolor{cvprblue}{rgb}{0.21,0.49,0.74}
\usepackage[pagebackref,breaklinks,colorlinks,allcolors=cvprblue]{hyperref}
\newcommand{\clip}{\operatorname{clip}}
\newcommand{\sg}{\operatorname{sg}}

\title{Evidence-Based Credit Assignment\\for Search-Augmented Language Models}
\author{%
  Wen Wang\textsuperscript{1,2}\thanks{Work done during an internship at Tencent.}\enspace
  Jindong Wang\textsuperscript{2}\thanks{Corresponding author.}\enspace
  Jiahua Bao\textsuperscript{3}\enspace
  Tu Yongsiqi\textsuperscript{4}\enspace
  Yihao Liu\textsuperscript{5}\enspace
  Junwei He\textsuperscript{1}\enspace
  Wenkui Fan\textsuperscript{6}\\[4pt]
  \small \textsuperscript{1}University of the Chinese Academy of Sciences\quad \textsuperscript{2}Tencent\quad \textsuperscript{3}Harbin Institute of Technology\\
  \small \textsuperscript{4}Renmin University of China\quad \textsuperscript{5}Peking University\quad \textsuperscript{6}Beijing Institute of Technology
}
\hypersetup{pdfauthor={Wen Wang, Jindong Wang, Jiahua Bao, Tu Yongsiqi, Yihao Liu, Junwei He, Wenkui Fan},pdftitle={Evidence-Based Credit Assignment for Search-Augmented Language Models}}
\input{sections/teaser_layout}
\begin{document}
\maketitle
\begin{abstract}
Reinforcement learning from terminal answer rewards enables search-augmented language models to interleave reasoning and retrieval without step-level annotations, yet outcome supervision assigns identical credit to every search in a trajectory regardless of whether it retrieved useful information. This limitation is especially severe when all sampled rollouts for a question fail, collapsing group-centered outcome advantages to zero even when trajectories gather markedly different evidence. We present \textbf{Evidence-based Credit Assignment (ECA)}, which augments terminal outcome rewards with a training-only auxiliary signal measuring how retrieved documents improve the likelihood of a reference answer. Rather than scoring the full dialogue prefix, ECA queries a frozen snapshot of the current policy under an evidence-only prompt containing solely the question and the deduplicated documents observed so far, isolating the utility of retrieved passages from surface variations in generated thoughts or search queries. Within this framework, we compare two return formulations, \textbf{ECA-Step} (which accumulates step-clipped potential increments) and \textbf{ECA-Horizon} (which preserves the telescoping horizon gain in potential), paired with cross-fold group normalization, novelty gating, and search-local token assignment. Across seven Search QA benchmarks, ECA-Horizon improves strict macro exact match over an outcome-only GRPO baseline by 1.03 and 4.80 percentage points on Qwen2.5-3B- and 7B-Instruct, respectively, outperforming both RLOO and GiGPO. Zero-shot evaluation on MultiHop-RAG further demonstrates strong transfer to multi-hop inference queries.
\end{abstract}
\input{sections/teaser}
\input{sections/introduction}
\input{sections/related_work}
\input{sections/method}
\input{sections/experiments}
\input{sections/results}
\input{sections/conclusion}
\input{sections/limitations}
{\small
\bibliographystyle{ieeenat_fullname}
\bibliography{references}}
\clearpage
\maketitlesupplementary
\label{supp:start}
\appendix
\setcounter{table}{0}
\setcounter{figure}{0}
\setcounter{equation}{0}
\renewcommand{\thetable}{S\arabic{table}}
\renewcommand{\thefigure}{S\arabic{figure}}
\renewcommand{\theequation}{S\arabic{equation}}
\renewcommand*{\theHequation}{S\arabic{equation}}
\input{sections/appendix}
\input{sections/related_comparison}
\input{sections/case_studies}
\end{document}

%% file: sections/teaser_layout.tex
\let\ECAOriginalIncludeGraphics\includegraphics
\renewcommand{\includegraphics}[2][]{%
  \ifstrequal{#2}{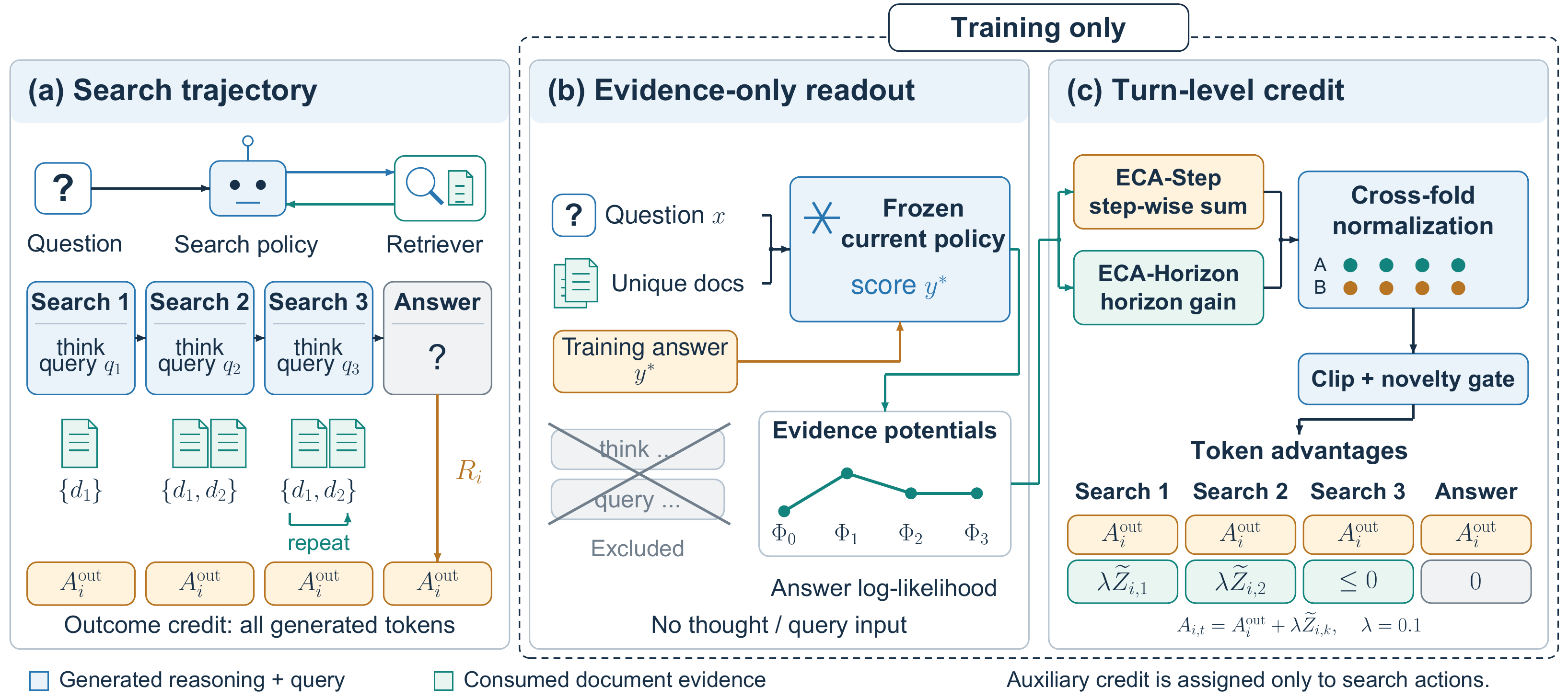}{%
    \ECAOriginalIncludeGraphics[width=0.85\textwidth]{#2}%
  }{%
    \ifstrequal{#2}{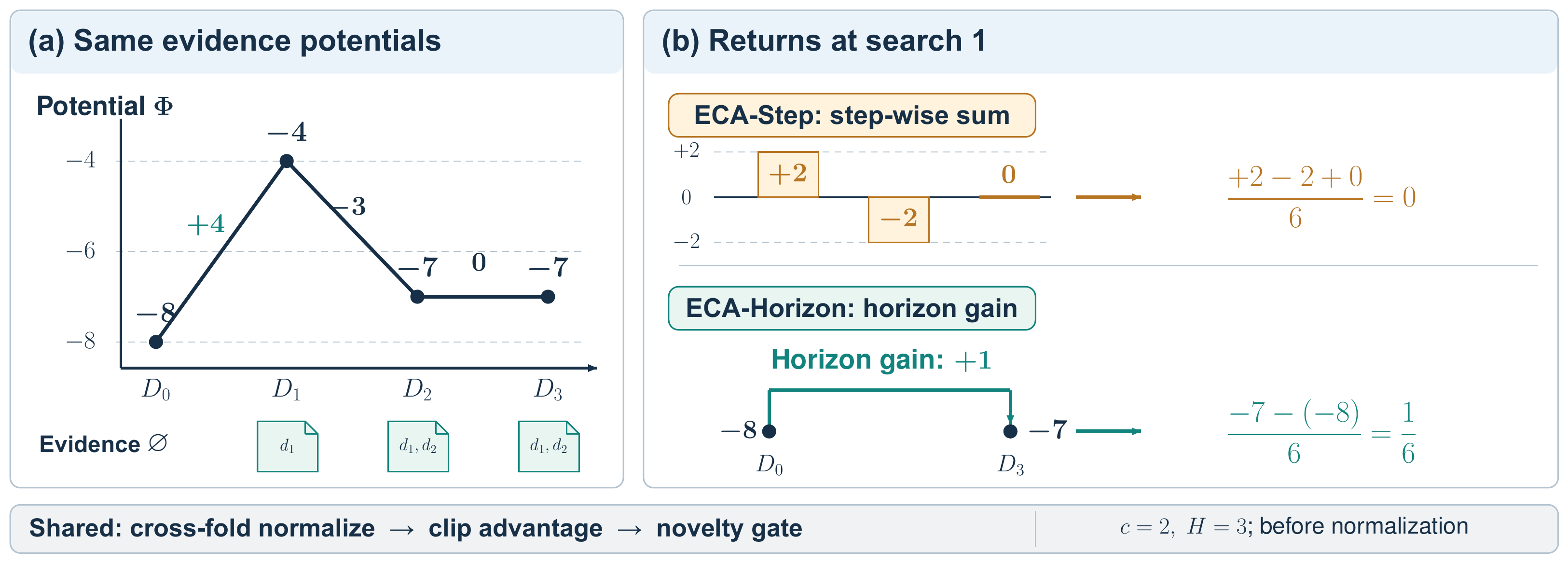}{%
      \ECAOriginalIncludeGraphics[width=0.75\textwidth]{#2}%
    }{%
      \ECAOriginalIncludeGraphics[#1]{#2}%
    }%
  }%
}

%% file: sections/teaser.tex
\begin{figure}[t]
\centering
\includegraphics[width=\linewidth]{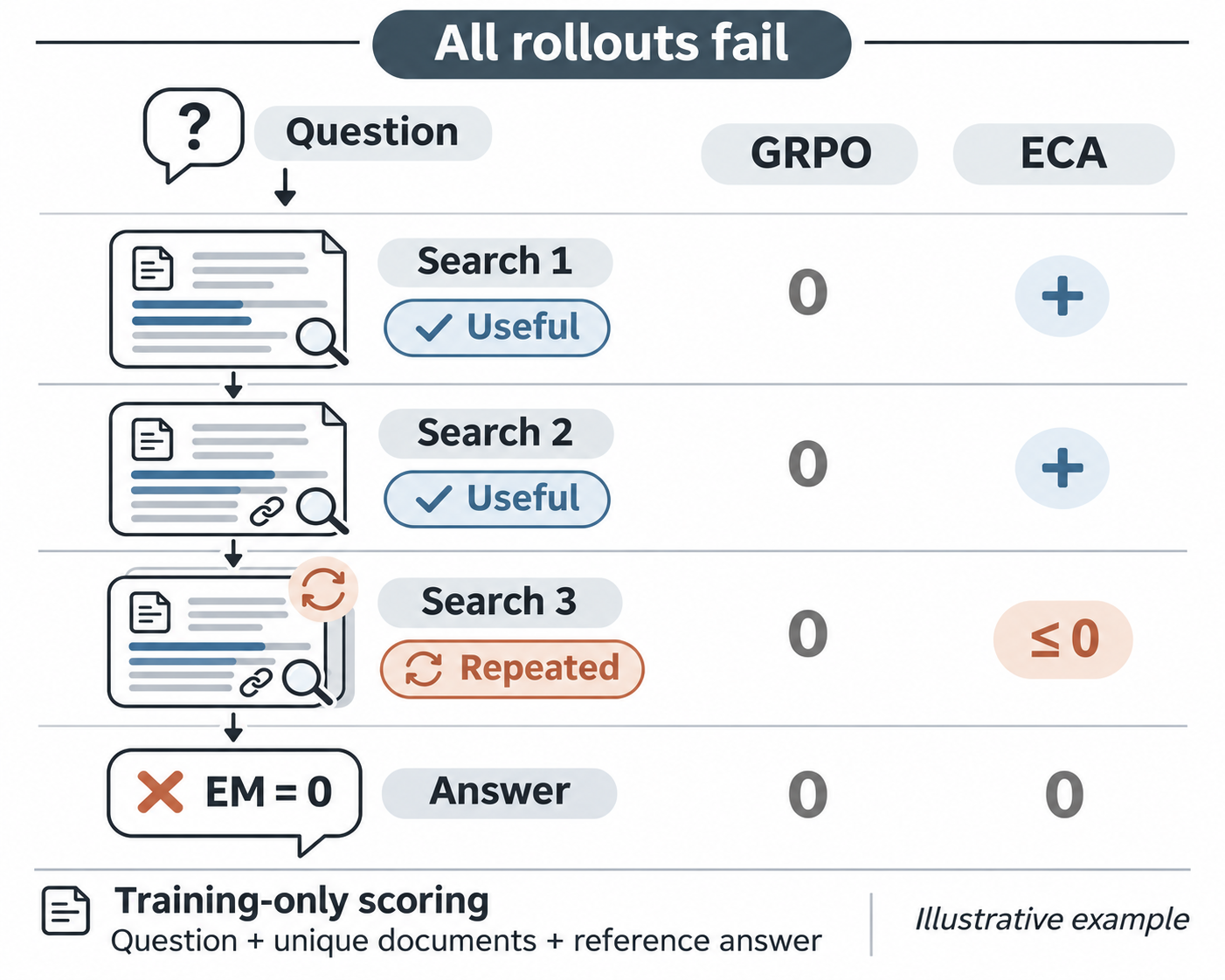}
\captionsetup{skip=3pt}
\caption{ECA can credit useful searches in all-failure groups while gating repeated evidence (illustrative example).}
\label{fig:teaser}
\end{figure}

%% file: sections/introduction.tex
\section{Introduction}
Search-augmented language models tackle knowledge-intensive questions by alternating between chain-of-thought reasoning, issuing search queries, and reading retrieved passages \citep{yao2023react}. Recent work has shown that reinforcement learning (RL) with terminal answer rewards can train these multi-turn interactions directly in retrieval environments \citep{jin2025searchr1,sun2025zerosearch,zheng-etal-2025-deepresearcher} using critic-free group-relative estimators such as GRPO and RLOO \citep{shao2024deepseekmath,ahmadian2024back}. Despite its simplicity, terminal outcome supervision provides a coarse signal for multi-step search: a single trajectory-level advantage is broadcast uniformly to every generated token. Consequently, a well-targeted search that uncovers a bridging document is penalized if the model later misstates the final answer, whereas a redundant query is reinforced whenever the model answers correctly from parametric memory or an earlier turn. This credit-assignment bottleneck becomes especially acute on hard questions where every sampled rollout in a group fails: centered outcome advantages collapse to zero across the group, discarding the gradient even when rollouts retrieved substantially different evidence.

Recent methods seek finer-grained supervision by grouping rollouts at repeated environment states \citep{feng2025gigpo}, scoring retrieved passages against supporting-document annotations \citep{zheng-etal-2025-stepsearch,huang2026retrievalcredit}, or shaping turn-level rewards via changes in the likelihood of the ground-truth answer \citep{wang2026igpo,xie2026tips}. When reference-answer likelihood is evaluated on the full dialogue history, however, the resulting potential reflects not only retrieved documents, but also the agent's intermediate reasoning and query phrasing. A policy can therefore alter its own potential merely by rewording a thought or echoing candidate answers, even when the underlying evidence set has not changed. This motivates a central question: \emph{can we assign effective step-level credit to search actions using an evidence-only potential that depends strictly on the question and retrieved documents?}

To address this question, we propose \textbf{Evidence-based Credit Assignment (ECA)}. During training, a frozen snapshot of the current policy teacher-forces a canonical reference answer under a dedicated scoring prompt containing only the question and the deduplicated documents consumed up to each search slot. Stripping generated thoughts and search queries from the scoring context makes the potential $\Phi(D)$ invariant to surface reasoning variations: a search action changes the potential if and only if it introduces new documents that shift the likelihood of the reference answer. This readout requires no gold passage labels, separate process reward models, or rollout continuations, and is discarded entirely at test time.

Within the ECA framework, we study how step-wise potential changes should be aggregated and normalized into token-level advantages. We compare two return constructions: \textbf{ECA-Step}, which clips single-step potential increments before accumulating future progress, and \textbf{ECA-Horizon}, which preserves the unclipped telescoping difference between the terminal horizon potential and the potential prior to a search. Both variants standardize returns via \emph{cross-fold normalization} (so a trajectory's own evidence gain does not inflate its normalization baseline) and apply a \emph{novelty gate} that withholds positive suffix credit when a query returns no new document IDs. Crucially, the resulting auxiliary advantage is added to the full search action (both the search call and its preceding reasoning tokens), leaving final-answer tokens under pure outcome supervision.

We evaluate ECA across seven Search QA benchmarks totaling 51,713 test questions with Qwen2.5-3B- and 7B-Instruct \citep{qwen2025report}, and test zero-shot transfer to MultiHop-RAG \citep{tang2024multihop}. Our main contributions are:
\begin{itemize}
\item We formulate an evidence-only answer potential and a bounded search-local advantage for outcome-supervised RL, restoring informative credit in over $97\%$ of all-failure rollout groups without passage annotations or external reward models.
\item We analyze how step-wise increment clipping interacts with multi-step evidence accumulation, comparing step-clipped returns (\textbf{ECA-Step}) and horizon-level telescoping returns (\textbf{ECA-Horizon}) alongside ablations on token placement and cross-fold normalization.
\item Across seven Search QA benchmarks, ECA-Horizon improves strict macro exact match over our matched GRPO baseline by $+1.03$ percentage points (pp) at 3B and $+4.80$ pp at 7B, outperforming RLOO and GiGPO at both scales while achieving strong zero-shot transfer on MultiHop-RAG inference questions.
\end{itemize}

%% file: sections/related_work.tex
\section{Related Work}
\paragraph{Credit assignment for multi-turn agents.}
Assigning step-level credit from sparse outcomes is a central challenge in RL for reasoning and tool-use agents. Process reward models supervise intermediate steps using human or automated annotations \citep{lightman2024verify,wang-etal-2024-math}, whereas VinePPO estimates step values via Monte Carlo rollout continuations \citep{kazemnejad2025vineppo}, and recent multi-turn formulations separate terminal, delayed, and per-turn rewards to reduce variance across long horizons \citep{li-etal-2026-turn,wei2026turncredit}. For retrieval agents, several works derive intermediate signals from corpus annotations: StepSearch rewards gold-document coverage while penalizing redundant queries \citep{zheng-etal-2025-stepsearch}, PiCA trains a reward model from pivot annotations \citep{liu2026pica}, and related methods ground credit in annotated retrieval dependencies \citep{huang2026retrievalcredit} or counterfactual query verification \citep{jiang2026igsd}. Without passage annotations, IGPO \citep{wang2026igpo} and TIPS \citep{xie2026tips} shape turn-level rewards using changes in reference-answer likelihood across the dialogue prefix, with IGPO normalizing immediate length-scaled log-likelihood gains before discounted accumulation. ECA builds on this likelihood-based paradigm, but restricts the scoring prompt strictly to the question and deduplicated retrieved documents (excluding generated thoughts and queries) and studies step-clipped versus horizon-level returns under search-local credit assignment.

\paragraph{Group-relative policy optimization.}
Critic-free estimators such as GRPO and RLOO compute relative advantages across multiple rollouts per prompt \citep{shao2024deepseekmath,ahmadian2024back}, a paradigm widely adopted in outcome-supervised search agents including Search-R1, ReSearch, and R1-Searcher \citep{jin2025searchr1,chen2025research,song2025r1searcher}. Because trajectory-level normalization assigns one scalar to all turns and collapses when all rollouts fail, recent extensions refine the comparison group or reweight round contributions: GiGPO groups actions at identical or similar intermediate states \citep{feng2025gigpo}, KnowGPO clusters rollouts by LLM-verified atomic knowledge conditions \citep{chen-etal-2026-knowgpo}, and CW-GRPO uses an LLM judge to reallocate outcome advantages across rounds \citep{wang2026cwgrpo}. Rather than requiring state collisions, external LLM verifiers, or successful donor trajectories, ECA retains the group-normalized terminal advantage and adds a bounded search-local residual standardized via cross-fold normalization within each question's rollout group (see Appendix~\ref{app:related-comparison} for a detailed comparison).

%% file: sections/method.tex
\section{Evidence-Based Credit Assignment}
\label{sec:method}
\begin{figure*}[t]
\centering
\includegraphics[width=\textwidth]{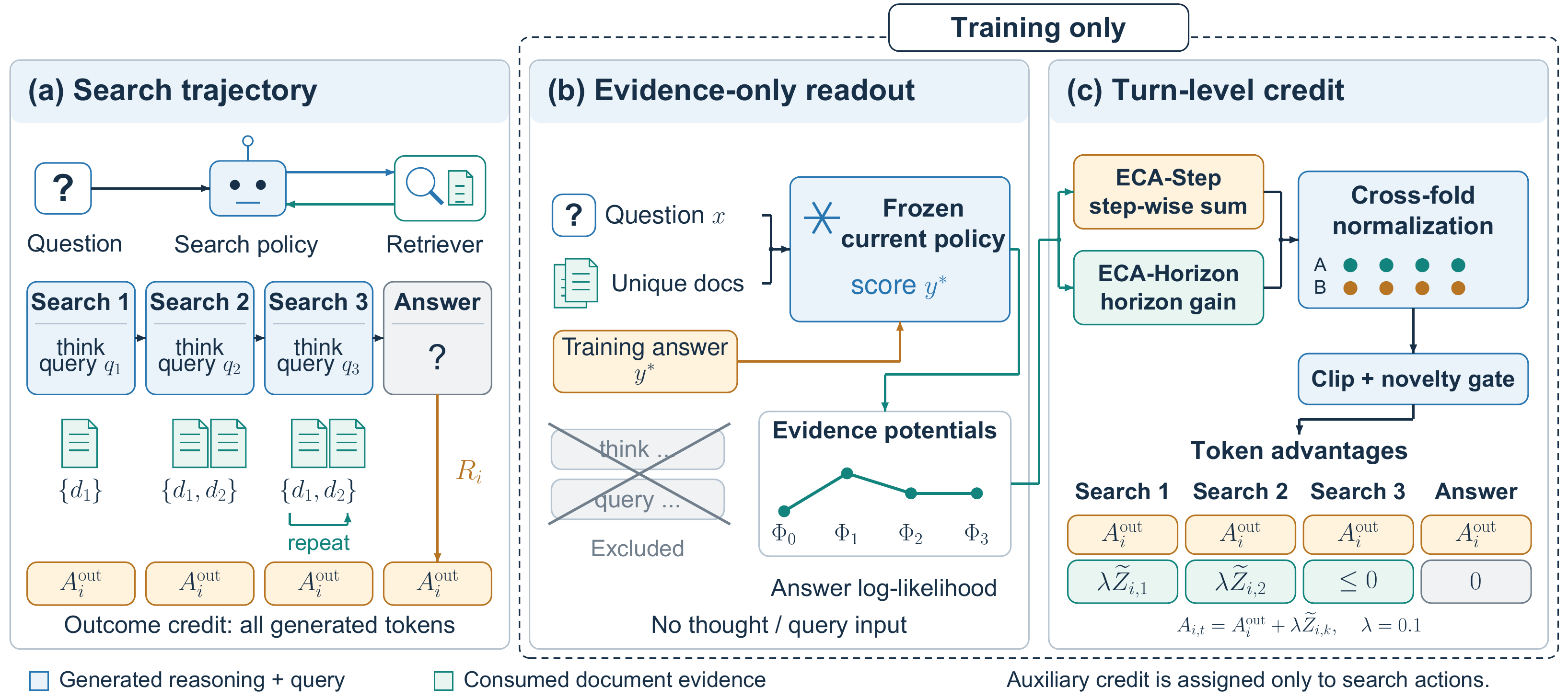}
\caption{Overview of Evidence-based Credit Assignment (ECA). \textbf{(a)} During rollout, the policy interleaves reasoning and search queries to accumulate documents, while terminal outcome credit $A^{\mathrm{out}}$ applies uniformly to all generated tokens. \textbf{(b)} During training, a frozen snapshot of the current policy scores the canonical reference answer $y^*$ given only the question $x$ and deduplicated documents $D$, excluding generated thoughts and queries. \textbf{(c)} Step-wise potentials are aggregated via ECA-Step or ECA-Horizon, standardized via cross-fold normalization, clipped, and passed through a novelty gate before being added to search-action tokens.}
\label{fig:framework}
\end{figure*}

\subsection{Search Trajectories and Outcome Supervision}
Given a question $x$, we sample a group of $G$ trajectories $\{\tau_i\}_{i=1}^G$ from the current behavior policy $\pi_{\theta_0}$. Each trajectory interleaves generated reasoning (\texttt{<think>}), search calls (\texttt{<search>}), retrieved document observations, and a terminal answer (\texttt{<answer>}). Let $R_i \in \{0,1\}$ denote the binary exact-match reward of the final answer. Following GRPO \citep{shao2024deepseekmath}, the trajectory-level outcome advantage is standardized across the group:
\begin{equation}
A_i^{\mathrm{out}}=\frac{R_i-\bar R}{s_R+\epsilon},
\label{eq:outcome}
\end{equation}
where $\bar R$ and $s_R$ denote the group mean and sample standard deviation of $\{R_j\}_{j=1}^G$, and $\epsilon=10^{-6}$. Retrieved observation tokens are masked out of the policy gradient loss. Whenever all $G$ trajectories answer incorrectly ($R_1 = \dots = R_G = 0$), $A_i^{\mathrm{out}}$ vanishes for every token in the group.

To align intermediate retrieval steps across trajectories of varying lengths, we index search slots over a fixed horizon $k \in \{1, \ldots, H\}$, where $H=3$ is the maximum search budget. Let $D_{i,k}$ denote the cumulative set of unique documents consumed by the policy after its $k$-th search, starting from $D_{i,0} = \varnothing$. If a trajectory terminates in fewer than $H$ searches, its remaining slots are padded by carrying forward its final document set ($D_{i,k} = D_{i,k-1}$) and receive no token-level credit. Defining returns over a fixed horizon $H$ avoids dividing cumulative gains by a variable, policy-chosen number of searches, which would otherwise reward premature termination after a single lucky retrieval.

\subsection{Evidence-Only Answer Potential}
\label{sec:potential}
A core principle of ECA is to measure the utility of retrieved documents without confounding it with how the agent phrased its intermediate thoughts or search queries. If a potential function conditions on the full dialogue prefix, the policy can inflate answer likelihood simply by hedging or hinting at candidate answers in its chain-of-thought, even when retrieval fails.

To isolate evidence progress, we evaluate reference-answer likelihood under an \emph{evidence-only} context (Figure~\ref{fig:framework}b). From the ground-truth answer aliases for question $x$, we select a canonical target $y^*$ with the fewest tokenizer tokens (breaking ties lexicographically) and denote its token length by $L$. Using a frozen snapshot of the current behavior policy $\pi_{\theta_0}$, we define the evidence potential of document set $D$ as the length-normalized log-likelihood of $y^*$ under teacher forcing:
\begin{equation}
\Phi_i(D)=\frac{1}{L}\sum_{\ell=1}^{L}
\log \pi_{\theta_0}\!\left(y^*_{\ell}\mid C(x,D),y^*_{<\ell}\right),
\label{eq:potential}
\end{equation}
where $C(x,D)$ is a structured prompt containing only the question $x$ and the deduplicated documents in $D$ sorted canonically by document ID. All generated thoughts, search queries, and sibling rollouts are excluded from $C(x,D)$, and gradients are stopped through $\Phi_i(D)$.

In practice, each document in $D$ is represented by its observed 1,024-character prefix within a 6,144-token scoring context (aborting scoring on context overflow so that evidence states are never silently altered by truncation). Because many rollout prefixes within a question group share identical document sets (including $D_{i,0}=\varnothing$ and repeated retrievals), we cache $\Phi_i(D)$ by document-ID set within each rollout batch. The scorer $\pi_{\theta_0}$ is held fixed within each rollout batch and synchronized with the updated actor before the next batch, requiring only batched forward passes during training and zero overhead at test time.

\subsection{Step-Wise and Horizon Potential Returns}
\begin{figure*}[t]
\centering
\includegraphics[width=\textwidth]{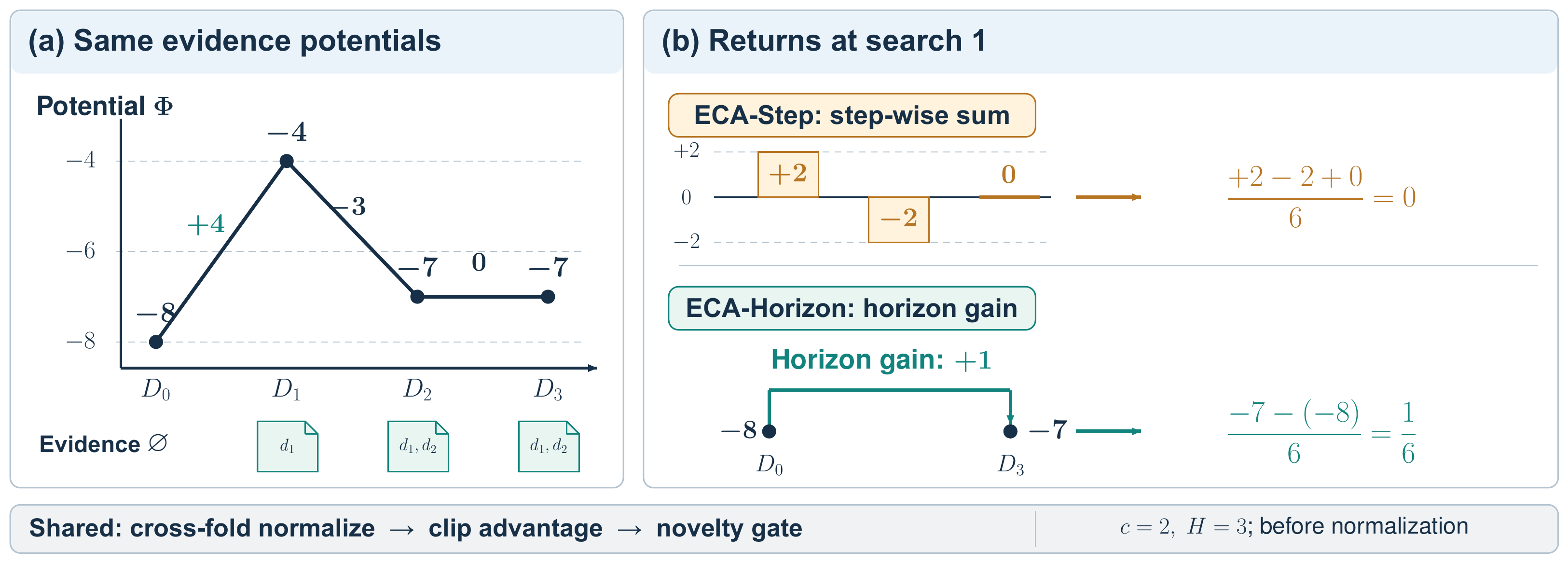}
\caption{Comparison of ECA-Step and ECA-Horizon return constructions on an illustrative trajectory with potentials $(-8,-4,-7,-7)$ and step-wise increments $(+4,-3,0)$ ($c=2, H=3$). \textbf{(a)} The first search raises the potential from $-8$ to $-4$, followed by a noisy second search ($-3$) and a redundant third search ($0$). \textbf{(b)} Clipping each step to $[-2,2]$ under ECA-Step saturates the $+4$ gain and $-3$ drop to $+2$ and $-2$, yielding zero return at search 1 ($U^{\mathrm S}_{i,1}=0$). ECA-Horizon preserves the horizon gain from $D_0$ to $D_3$, yielding $U^{\mathrm H}_{i,1}=(-7-(-8))/6=1/6$ before the shared normalization and gating stages.}
\label{fig:variants}
\end{figure*}

Let $\Delta_{i,k}$ denote the single-step change in evidence potential produced by the $k$-th search:
\begin{equation}
\Delta_{i,k}=\Phi_i(D_{i,k})-\Phi_i(D_{i,k-1}).
\end{equation}
Because a search at slot $k$ can unlock follow-up queries in slots $j > k$, we evaluate slot $k$ by the cumulative evidence progress from $k$ through $H$. We compare two return constructions prior to normalization and gating (Figure~\ref{fig:variants}).

\paragraph{ECA-Step (step-clipped return).}
Log-likelihood increments $\Delta_{i,j}$ can exhibit sharp spikes when a passage directly names the target entity or introduces strong distractors. To cap the influence of any single step, ECA-Step clips each increment to $[-c, c]$ (with $c=2$) before summing over the remaining steps:
\begin{equation}
U^{\mathrm S}_{i,k}=\frac{1}{cH}\sum_{j=k}^{H}
\clip(\Delta_{i,j},-c,c).
\label{eq:v1}
\end{equation}
Scaling by $cH$ bounds $|U^{\mathrm S}_{i,k}| \le (H-k+1)/H$, though clipping before summation breaks telescoping whenever positive and negative steps saturate asymmetrically.

\paragraph{ECA-Horizon (horizon-gain return).}
To preserve the exact cumulative change in evidence potential across the remaining horizon, ECA-Horizon sums the unclipped increments from slot $k$ to $H$, telescoping directly to the difference between the terminal horizon potential and the potential immediately before search $k$:
\begin{align}
U^{\mathrm H}_{i,k}
&=\frac{1}{cH}\sum_{j=k}^{H}\Delta_{i,j}\nonumber\\
&=\frac{\Phi_i(D_{i,H})-\Phi_i(D_{i,k-1})}{cH}.
\label{eq:v2}
\end{align}
While $U^{\mathrm H}_{i,k}$ preserves exact horizon progress regardless of intermediate spikes, stability is maintained by the shared post-normalization clipping stage below.

When all increments in a group satisfy $|\Delta_{i,j}| \le c$, ECA-Step and ECA-Horizon produce identical returns. When increments exceed $c$, however, step-wise clipping can alter both the magnitude and ranking of search actions. As shown in Figure~\ref{fig:variants}, if three searches yield potential increments $(+4, -3, 0)$ with $c=2$ and $H=3$, the trajectory finishes with a horizon gain of $+1$, giving $U^{\mathrm H}_{i,1} = 1/6$ under ECA-Horizon. Under ECA-Step, the $+4$ gain and $-3$ drop saturate to $+2$ and $-2$, cancelling out to $U^{\mathrm S}_{i,1} = (2 - 2 + 0)/6 = 0$ and erasing credit for the productive first search.

\subsection{Cross-Fold Normalization and Novelty Gating}
\label{sec:normalization}
To convert raw returns $U^m_{i,k}$ ($m \in \{\mathrm S, \mathrm H\}$) into well-scaled advantages across questions of varying difficulty, we standardize returns within each question group. Unlike binary terminal rewards $R_i \in \{0,1\}$, intermediate evidence returns are continuous and heavy-tailed: on a hard question where only one of $G=8$ rollouts discovers a high-value passage ($u > 0$) while the other seven yield $0$, normalizing across all eight rollouts includes the spike $u$ in its own reference mean ($u/G$) and standard deviation ($s = u/\sqrt{G}$), causing the magnitude $u$ to cancel out of $(u - u/G)/s$ once $s \ge \sigma_{\min}$ (Appendix~\ref{app:properties}).

To keep baseline statistics independent of the evaluated trajectory while training on all $G$ rollouts, we randomly partition the $G=8$ trajectories of each question into two disjoint folds of four (independently of rewards) and standardize each trajectory against the mean $\mu_{-f(i),k}^{m}$ and sample standard deviation $s_{-f(i),k}^{m}$ of the \emph{opposite} fold:
\begin{equation}
Z^{m}_{i,k}=\clip\!\left(
\frac{U^{m}_{i,k}-\mu_{-f(i),k}^{m}}{\max(s_{-f(i),k}^{m},\sigma_{\min})},-2,2\right),
\label{eq:norm}
\end{equation}
where $\sigma_{\min}=0.25$ prevents division by near-zero dispersion when reference rollouts have similar potentials, and clipping to $[-2,2]$ bounds outliers. Because fold statistics do not filter by terminal answer correctness, informative searches in failed trajectories can still receive positive credit. We omit $m$ below as the remaining steps are shared.

Finally, because $U_{i,k}$ accumulates suffix progress from slot $k$ through $H$, a search at slot $k$ that retrieves only already-seen documents ($D_{i,k}=D_{i,k-1}$, so $\Delta_{i,k}=0$) could otherwise inherit a positive return from a productive search at slot $k+1$. To prevent redundant queries from free-riding on subsequent discoveries, we apply a \emph{novelty gate}:
\begin{equation}
\widetilde Z_{i,k}=\begin{cases}
\min(0,Z_{i,k}),&D_{i,k}=D_{i,k-1},\\
Z_{i,k},&\text{otherwise}.
\end{cases}
\label{eq:gate}
\end{equation}
Equation~\ref{eq:gate} zeroes out positive auxiliary credit whenever a search introduces no novel document IDs, while preserving negative penalties (when a wasted search leaves the rollout behind its peers) and the trajectory-level outcome advantage $A_i^{\mathrm{out}}$.

\subsection{Search-Local Policy Update}
An effective search requires not only emitting the query string inside \texttt{<search>}, but also formulating the information need in the preceding \texttt{<think>} block. Let $M_{i,k,t} \in \{0,1\}$ indicate whether generated policy token $t$ belongs to the $k$-th executed and consumed search action, spanning both its reasoning tokens and its search-call tokens while excluding retrieved observations and the final-answer action. We combine the global outcome advantage and local evidence residual as:
\begin{equation}
A_{i,t}=A_i^{\mathrm{out}}+
\lambda\sum_{k=1}^{H}M_{i,k,t}\widetilde Z_{i,k},
\qquad \lambda=0.1,
\label{eq:combined}
\end{equation}
so that the auxiliary term adjusts search-action advantages within $[-0.2, 0.2]$ while final-answer tokens are updated solely by $A_i^{\mathrm{out}}$. With probability ratio $\rho_{i,t}(\theta)=\pi_\theta(a_{i,t}\mid h_{i,t})/\pi_{\theta_0}(a_{i,t}\mid h_{i,t})$ over generated policy tokens $\mathcal P_i$, we perform a single fresh-batch policy gradient update per rollout batch of $B$ trajectories:
\begin{equation}
\mathcal L(\theta)=-\frac{1}{B}\sum_{i=1}^{B}
\sum_{t\in\mathcal P_i}
\rho_{i,t}(\theta)
\sg(A_{i,t}),
\label{eq:loss}
\end{equation}
without ratio clipping, KL penalties, or entropy regularization. Because our GRPO baseline optimizes Eq.~\ref{eq:loss} with $\lambda=0$, comparing ECA against GRPO directly isolates the effect of the search-local evidence residual $\lambda \widetilde Z_{i,k}$. In all-failure rollout groups where $A_i^{\mathrm{out}}=0$, variation in $\widetilde Z_{i,k}$ continues to provide a structured learning signal for search actions (formal properties in Appendix~\ref{app:properties}).

%% file: sections/experiments.tex
\section{Experimental Setup}
\label{sec:setup}
\paragraph{Models and training protocol.}
We conduct experiments on Qwen2.5-3B-Instruct and Qwen2.5-7B-Instruct. All RL methods are trained on an 8,192-question pool (4,096 from Natural Questions and 4,096 from HotpotQA), with an independent 1,024-question in-domain development set. Training runs for 256 optimizer updates with learning rate $10^{-6}$, sampling 32 questions with $G=8$ rollouts per question at each update (256 trajectories per step; 65,536 total). Forward passes use BF16 mixed precision with FP32 master weights and AdamW states (with CPU optimizer offloading at 7B).

\paragraph{Search environment.}
Within each episode, the agent can take up to four actions (at most $H=3$ search calls followed by a final answer), generating up to 512 tokens per action within a 4,096-token context window. Rollouts are sampled at temperature 1.0 during training and decoded greedily at evaluation. Retrieval is served by a frozen E5-base-v2 retriever \citep{wang2022e5} over Wikipedia18, returning the top-3 passages per query via exact FP32 inner-product search. Malformed or unfinished episodes are retained in evaluation and scored as incorrect if they fail to produce a matching answer.

\paragraph{Baselines and checkpoint selection.}
We compare ECA-Step and ECA-Horizon against three RL baselines reproduced under the same 8,192-question training pool and 65,536-trajectory budget: (i) \textbf{GRPO} ($\lambda=0$ in Eq.~\ref{eq:combined}), (ii) \textbf{RLOO} \citep{ahmadian2024back}, and (iii) \textbf{GiGPO (w/std)} \citep{feng2025gigpo}, alongside the untrained \textbf{Base agent}. We also report published results for R1-Instruct, Search-R1, ZeroSearch, and StepSearch from \citet{feng2025gigpo}, and TIPS from \citet{xie2026tips}. Checkpoints are evaluated every 32 updates on the 1,024-question development set (breaking ties toward earlier steps), and the single best development checkpoint per run is used across all test suites and transfer evaluations. Both ECA variants use $\lambda=0.1$, selected on the development set for ECA-Step and transferred directly to ECA-Horizon without retuning (Appendix~\ref{app:protocol}).

\paragraph{Evaluation benchmarks and metrics.}
Following Search-R1 and GiGPO \citep{jin2025searchr1,feng2025gigpo}, we evaluate on seven QA benchmarks totaling 51,713 test questions: two in-domain datasets (NQ, HotpotQA) and five out-of-domain datasets (TriviaQA, PopQA, 2WikiMultiHopQA, MuSiQue, and Bamboogle). Our primary metric is strict question-level Exact Match (EM). For question $q$, let $e_q = 1$ if the normalized non-empty prediction matches any normalized ground-truth alias (after lowercasing and stripping punctuation, articles, and extra whitespace), and $e_q = 0$ otherwise. We report the unweighted macro-average across the seven datasets:
\begin{equation}
\mathrm{EM}=\frac{100}{7}\sum_{s=1}^{7}\frac{1}{|\mathcal Q_s|}
\sum_{q\in\mathcal Q_s}e_q,
\label{eq:score}
\end{equation}
where $\mathcal Q_s$ is the test set of benchmark $s$ (Appendix~\ref{app:scoring}).

\paragraph{Zero-shot transfer to MultiHop-RAG.}
To test whether learned search behaviors generalize to a new corpus and longer documents without further tuning, we evaluate the selected Search QA checkpoints zero-shot on MultiHop-RAG \citep{tang2024multihop}. The agent retrieves top-3 chunks (using E5-base-v2 and up to three searches within a 16,384-token context) over 6,532 chunks segmented from 609 news articles. We report performance on the 816 multi-hop inference questions following the TreeHop protocol \citep{li2026treehop}.

%% file: sections/results.tex
\section{Results and Analysis}
\input{tables/search_main}

\subsection{Main Results on Search QA}
Table~\ref{tab:search} presents strict question-level EM across the seven Search QA benchmarks. Starting from the untrained Base agents ($21.75$ macro EM at 3B and $29.93$ at 7B), outcome-supervised RL substantially improves multi-turn retrieval and answering, with GRPO reaching $37.19$ (3B) and $37.82$ (7B). Augmenting terminal outcome supervision with our evidence-only credit signal yields consistent further gains across both model scales. At 3B, \textbf{ECA-Step} and \textbf{ECA-Horizon} reach $38.05$ and $38.22$ macro EM, outperforming GRPO by $+0.86$ and $+1.03$ percentage points (pp) and improving over the Base agent by $+16.29$ and $+16.46$ pp, respectively. At 7B, the advantage of evidence-based credit is even more pronounced: \textbf{ECA-Step} reaches $41.65$ EM ($+3.83$ pp over GRPO) and \textbf{ECA-Horizon} achieves $42.62$ EM, surpassing GRPO by $+4.80$ pp, RLOO ($40.56$) by $+2.06$ pp, and the Base agent by $+12.69$ pp.

ECA-Horizon also outperforms the state-grouped GiGPO (w/std) baseline at both scales ($38.22$ vs.\ $36.96$ at 3B, $+1.26$ pp; $42.62$ vs.\ $41.09$ at 7B, $+1.53$ pp). Compared against published likelihood-shaping and search RL baselines, ECA-Horizon exceeds TIPS ($33.60$ at 3B and $41.71$ at 7B \citep{xie2026tips}) by $+4.62$ and $+0.91$ pp and Search-R1 ($32.5$ and $38.5$) by $+5.72$ and $+4.12$ pp.

Examining individual benchmarks reveals two notable trends. First, the benefit of ECA scales strongly with model capacity: moving from 3B to 7B widens ECA-Horizon's margin over GRPO from $+1.03$ pp to $+4.80$ pp, with large gains on HotpotQA ($+4.09$ pp), Bamboogle ($+8.00$ pp), and NQ ($+5.57$ pp). Because ECA uses the policy itself as the frozen scorer $\Phi_i(D)$, a stronger 7B backbone provides sharper answer-likelihood estimates while avoiding GRPO's early plateau on multi-hop retrieval. Second, preserving horizon-level gains (\textbf{ECA-Horizon}) outperforms step-wise clipping (\textbf{ECA-Step}) on aggregate EM at both scales, especially on 7B 2WikiMultiHopQA ($39.35$ vs.\ $35.95$) and Bamboogle ($40.00$ vs.\ $37.60$), while ECA-Step remains competitive on splits such as 7B MuSiQue ($17.75$ vs.\ $16.88$).

\subsection{Ablations on Credit Assignment Design}
\input{tables/ablation}
Table~\ref{tab:ablation} isolates the core design choices of ECA at 3B ($\lambda=0.1$). First, \textbf{where} auxiliary credit is applied along the trajectory is critical. Restricting ECA-Step's credit mask to the \texttt{<search>} query tokens alone (excluding the preceding \texttt{<think>} tokens) causes macro EM to drop from $38.05$ to $35.27$, falling below outcome-only GRPO ($37.19$). Conversely, broadcasting the first-slot evidence advantage across all trajectory tokens (without local novelty gating) reaches only $36.42$ EM. Deciding \emph{what} to query happens during the reasoning tokens preceding the search call: supervising only the query tag decouples search planning from query execution, whereas trajectory-wide broadcasting blurs turn-level attribution and leaks search credit into the final answer step.

Second, comparing \textbf{return constructions} at 3B, both ECA-Step ($38.05$ EM, $46.76$ F1) and ECA-Horizon ($38.22$ EM, $45.80$ F1) substantially outperform outcome-only GRPO ($37.19$ EM) as well as the query-only ($35.27$ EM) and broadcast ($36.42$ EM) ablations (see Appendix~\ref{app:properties} for the analysis of cross-fold vs.\ full-group normalization). Between the two variants, ECA-Step attains slightly higher token F1 at 3B, whereas ECA-Horizon achieves higher strict EM at both 3B ($+0.17$ pp) and 7B ($+0.97$ pp), confirming that preserving unclipped horizon gains becomes increasingly beneficial as model capacity scales.

\subsection{Signal Recovery in All-Failure Groups and Agent Behavior}
\begin{figure*}[t]
\centering
\includegraphics[width=\linewidth]{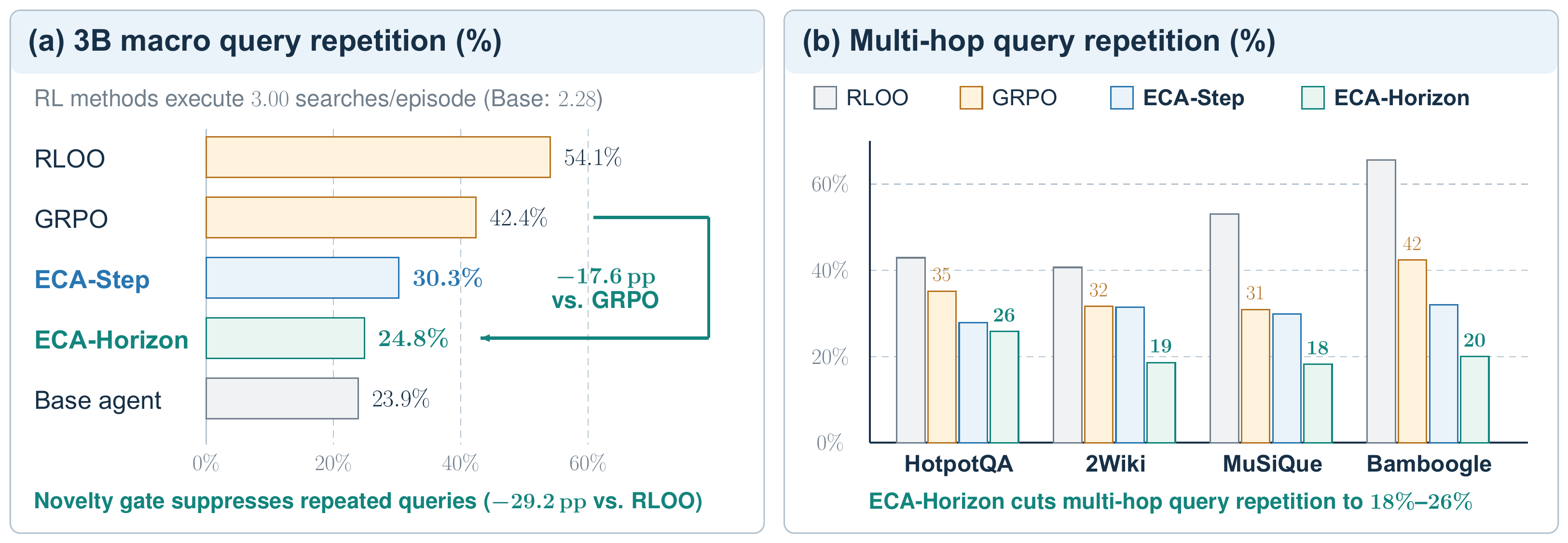}
\caption{Episode-level duplicate search query rate (\%) on Qwen2.5-3B-Instruct across (a) the seven-dataset macro average and (b) individual multi-hop QA benchmarks. All RL-trained policies execute the full budget of $3.00$ searches per episode (vs.\ $2.28$ for the Base agent); pairing search-local evidence credit with novelty gating suppresses the redundant repeat queries induced by outcome-only RL.}
\label{fig:repetition}
\end{figure*}
A primary motivation for ECA is recovering informative gradients when every rollout for a question fails terminal EM ($A_i^{\mathrm{out}} = 0$). Across the 256 training updates at 3B, ECA-Horizon produces nonzero auxiliary advantages in $3,213$ of $3,285$ all-failure question groups ($97.8\%$). Crucially, however, simply injecting nonzero variance into failed groups is insufficient on its own: the query-only ablation in Table~\ref{tab:ablation} also generates nonzero signals in $3,371$ of $3,441$ all-failure groups ($98.0\%$), yet degrades test accuracy below GRPO. Dense evidence signals translate into stronger policies only when paired with search-local credit placement that reinforces the reasoning behind productive queries.

Examining agent interaction patterns across all 51,713 test questions (Figure~\ref{fig:repetition} and Appendix~\ref{app:behavior}, Tables~\ref{tab:behavior}--\ref{tab:paired-behavior}) reveals \emph{how} ECA improves retrieval quality. While the untrained Base models average $3.36$ (3B) and $2.66$ (7B) actions per episode, all RL-trained methods converge to the full budget of $4.00$ actions and $3.00$ searches while virtually eliminating malformed actions. At 3B, however, outcome-only RL frequently pads this three-search budget by repeating earlier queries (Figure~\ref{fig:repetition}): the fraction of episodes containing duplicate search queries rises from $23.86\%$ in the Base agent to $42.40\%$ under GRPO and $54.07\%$ under RLOO. Because ECA pairs step-local evidence credit with novelty gating that withholds positive credit from redundant retrievals ($D_{i,k}=D_{i,k-1}$), it cuts 3B query repetition down to $30.26\%$ (ECA-Step) and $24.84\%$ (ECA-Horizon), converting wasted repeat searches into diverse follow-up queries across multi-hop benchmarks while generating fewer policy tokens ($200.9$ for ECA-Horizon vs.\ $257.1$ for ECA-Step). On a paired per-question basis against the Base agent, ECA-Horizon converts $12,340$ previously wrong answers to correct at 3B (vs.\ $2,166$ regressions) and $8,861$ to correct at 7B (vs.\ $2,158$ regressions).

\subsection{Zero-Shot Transfer to MultiHop-RAG}
\input{tables/inference816}
Table~\ref{tab:inference} reports zero-shot transfer to the 816 multi-hop inference questions of MultiHop-RAG \citep{tang2024multihop,li2026treehop}. On these compositional inference queries, where the agent must synthesize facts across multiple news articles, both ECA variants outperform all baselines: at 3B, ECA-Step reaches $84.68$ EM ($+4.66$ pp over GRPO's $80.02$ and $+1.22$ pp over RLOO's $83.46$), while at 7B, ECA-Horizon reaches $89.22$ EM ($+6.38$ pp over GRPO's $82.84$, $+1.97$ pp over GiGPO's $87.25$, and $+1.11$ pp over RLOO's $88.11$).

%% file: tables/search_main.tex
\begin{table*}[t]
\caption{Question-level exact match (EM, \%) across the seven-dataset Search QA benchmark, with macro average (Avg., Eq.~\ref{eq:score}). $\dagger$ denotes published reference results from \citet{feng2025gigpo} (Table~2) and \citet{xie2026tips} (Table~1). Best and second-best scores per column within each model scale are shown in \textbf{bold} and \underline{underlined}.}
\label{tab:search}
\centering\small
\setlength{\tabcolsep}{4pt}
\begin{tabular}{lrrrrrrrr}
\toprule
Method & NQ & TriviaQA & PopQA & Hotpot & 2Wiki & MuSiQue & Bamboogle & Avg. \\
\midrule
\multicolumn{9}{l}{\textbf{Qwen2.5-3B-Instruct}} \\
R1-Instruct$^{\dagger}$ & 27.0 & 53.7 & 19.9 & 23.7 & 29.2 & 7.2 & 29.3 & 27.1 \\
Search-R1$^{\dagger}$ & 34.1 & 54.5 & 37.8 & 32.4 & 31.9 & 10.3 & 26.4 & 32.5 \\
ZeroSearch$^{\dagger}$ & 41.4 & 57.4 & \underline{44.8} & 27.4 & 30.0 & 9.8 & 11.1 & 31.7 \\
StepSearch$^{\dagger}$ & -- & -- & -- & 34.5 & 32.0 & \textbf{17.4} & \textbf{34.4} & -- \\
TIPS$^{\dagger}$ & 43.46 & 58.80 & 42.76 & 31.40 & 29.25 & 8.73 & 20.80 & 33.60 \\
\cmidrule(lr){1-9}
Base agent & 26.20 & 40.06 & 23.53 & 19.64 & 14.33 & 6.12 & 22.40 & 21.75 \\
GRPO & 41.16 & 58.38 & 41.24 & \underline{38.53} & 37.10 & \underline{14.32} & 29.60 & 37.19 \\
RLOO & 42.55 & 59.68 & 41.43 & 35.60 & 33.41 & 11.34 & 24.00 & 35.43 \\
GiGPO (w/std) & \underline{44.63} & \textbf{61.19} & \textbf{45.49} & 33.98 & 34.54 & 11.67 & 27.20 & 36.96 \\
ECA-Step & 42.24 & \underline{59.74} & 42.01 & 38.38 & \textbf{38.26} & 13.69 & \underline{32.00} & \underline{38.05} \\
ECA-Horizon & \textbf{44.71} & 59.49 & 43.13 & \textbf{39.12} & \underline{37.25} & 13.41 & 30.40 & \textbf{38.22} \\
\midrule
\multicolumn{9}{l}{\textbf{Qwen2.5-7B-Instruct}} \\
R1-Instruct$^{\dagger}$ & 21.0 & 44.9 & 17.1 & 20.8 & 27.5 & 6.0 & 19.2 & 22.4 \\
Search-R1$^{\dagger}$ & 39.3 & 61.0 & 39.7 & 37.0 & 40.1 & 14.6 & 36.8 & 38.5 \\
ZeroSearch$^{\dagger}$ & 43.6 & 61.8 & \textbf{51.5} & 34.6 & 35.2 & \underline{18.4} & 27.8 & 39.1 \\
StepSearch$^{\dagger}$ & -- & -- & -- & 38.6 & 36.6 & \textbf{22.6} & \textbf{40.0} & -- \\
TIPS$^{\dagger}$ & 43.38 & \underline{64.31} & 44.52 & \underline{42.95} & \textbf{42.96} & 17.05 & 36.80 & \underline{41.71} \\
\cmidrule(lr){1-9}
Base agent & 33.55 & 54.84 & 36.36 & 27.09 & 21.45 & 10.63 & 25.60 & 29.93 \\
GRPO & 42.05 & 59.92 & 41.58 & 38.96 & 36.74 & 13.49 & 32.00 & 37.82 \\
RLOO & 45.93 & 63.38 & \underline{46.87} & 40.23 & 37.88 & 15.27 & 34.40 & 40.56 \\
GiGPO (w/std) & 44.46 & 62.58 & 45.15 & 38.54 & \underline{41.21} & 16.51 & \underline{39.20} & 41.09 \\
ECA-Step & \underline{47.06} & 63.90 & 46.80 & 42.48 & 35.95 & 17.75 & 37.60 & 41.65 \\
ECA-Horizon & \textbf{47.62} & \textbf{64.72} & 46.70 & \textbf{43.05} & 39.35 & 16.88 & \textbf{40.00} & \textbf{42.62} \\
\bottomrule
\end{tabular}
\end{table*}

%% file: tables/ablation.tex
\begin{table}[t]
\caption{Credit-assignment ablations on Qwen2.5-3B-Instruct ($\lambda=0.1$). EM and token F1 are macro averages (\%) across the seven Search QA datasets. Best and second-best scores in each column are shown in \textbf{bold} and \underline{underlined}.}
\label{tab:ablation}
\centering\small
\begin{tabular}{lrr}
\toprule 3B variant & EM & F1 \\ \midrule
GRPO & 37.19 & \underline{45.87} \\
ECA-Step & \underline{38.05} & \textbf{46.76} \\
ECA-Horizon & \textbf{38.22} & 45.80 \\
ECA-Step (query only) & 35.27 & 43.56 \\
ECA-Step (broadcast) & 36.42 & 44.96 \\
\bottomrule\end{tabular}
\end{table}

%% file: tables/inference816.tex
\begin{table}[t]
\caption{Zero-shot transfer on MultiHop-RAG Inference-816 (EM, \%), evaluated under the shared search agent and TreeHop scoring protocol. Best and second-best scores in each column are shown in \textbf{bold} and \underline{underlined}.}
\label{tab:inference}
\centering\small
\begin{tabular}{lrr}
\toprule Method & 3B & 7B \\ \midrule
Base & 59.93 & 76.84 \\
GRPO & 80.02 & 82.84 \\
RLOO & \underline{83.46} & 88.11 \\
GiGPO (w/std) & 65.93 & 87.25 \\
ECA-Step & \textbf{84.68} & \underline{88.85} \\
ECA-Horizon & 82.48 & \textbf{89.22} \\
\bottomrule\end{tabular}
\end{table}

%% file: sections/conclusion.tex
\section{Conclusion}
We presented Evidence-based Credit Assignment (ECA), a training-only credit assignment framework for search-augmented language models. By evaluating the log-likelihood of a canonical reference answer under an evidence-only prompt containing solely the question and deduplicated retrieved documents, ECA isolates the informational contribution of each search action from surface variations in intermediate reasoning and query phrasing. Comparing step-clipped increments (\textbf{ECA-Step}) against telescoping horizon potential differences (\textbf{ECA-Horizon}) within a shared training pipeline, we showed that ECA-Horizon consistently improves strict macro exact match over outcome-only GRPO, RLOO, and state-grouped GiGPO across seven Search QA benchmarks at both 3B and 7B scales, and transfers effectively zero-shot to MultiHop-RAG multi-hop inference queries.

%% file: sections/limitations.tex
\section*{Limitations}
Several limitations of our study point to directions for future work. First, our current empirical evaluation focuses on the search-augmented question answering domain and has not yet been extended to broader multi-turn agent settings such as code generation, interactive debugging, or general tool-use environments. Adapting evidence-only credit assignment to these broader agent domains (where intermediate observations take the form of execution traces, compiler diagnostics, or structured API responses) is an important direction for future work. In addition, computing $\Phi_i(D)$ introduces batched teacher-forced forward passes during training, although test-time inference cost remains unchanged.

Second, reference-answer log-likelihood under a single canonical alias $y^*$ is an imperfect proxy for document utility: a passage may provide an intermediate bridge entity without immediately raising the probability of final answer tokens, and document-ID deduplication does not detect near-duplicate text across different articles.
\label{main:content-end}

%% file: sections/appendix.tex
\input{sections/behavior}

\section{Implementation and Reproducibility Details}
\label{app:protocol}
\subsection{Training and Checkpoint Configuration}
\begin{table}[t]
\caption{Shared training and evaluation configuration for ECA. Zero-shot transfer on MultiHop-RAG uses a 16,384-token policy context.}
\label{tab:config}
\centering\small
\begin{tabular}{lr}
\toprule Hyperparameter & Value \\ \midrule
Training pool / development set & 8,192 / 1,024 \\
Questions per update / rollout group $G$ & 32 / 8 \\
Total optimizer updates & 256 \\
Optimizer / learning rate & AdamW / $10^{-6}$ \\
Max actions / search budget $H$ & 4 / 3 \\
Max generated tokens per action & 512 \\
Policy context limit (Search QA) & 4,096 \\
Evidence scorer context limit & 6,144 \\
Increment clip $c$ / std floor $\sigma_{\min}$ & 2 / 0.25 \\
Normalized advantage clip & $[-2,2]$ \\
Auxiliary coefficient $\lambda$ & 0.1 \\
Checkpoint evaluation interval & 32 updates \\
\bottomrule
\end{tabular}
\end{table}

Table~\ref{tab:config} lists the shared hyperparameters for ECA-Step and ECA-Horizon. Every run trains for 256 updates and is evaluated every 32 updates on the 1,024-question in-domain development set, selecting the single best development checkpoint per run for all test evaluations. For memory efficiency, the shared trainer processes microbatches of 8 trajectories (16,384-token padded budget) at 3B and 4 trajectories (8,192-token budget with CPU AdamW state offloading) at 7B, accumulating gradients over all 256 trajectories per update before taking a single optimizer step.

For our reproduced GiGPO baselines, we follow \citet{feng2025gigpo} with discount factor $\gamma=0.95$, step-advantage weight $1.0$, and state-similarity threshold $0.9$ under the same 8,192-question training pool and 65,536-trajectory budget.

\subsection{Step-by-Step ECA Training Loop}
Each training update proceeds as follows:
\begin{enumerate}
\item \textbf{Rollout generation}: Sample $G=8$ trajectories for each of the 32 training questions using $\pi_{\theta_0}$ at temperature 1.0, and compute terminal exact-match rewards $R_i \in \{0,1\}$ and outcome advantages $A_i^{\mathrm{out}}$ (Eq.~\ref{eq:outcome}).
\item \textbf{Evidence state reconstruction}: For each trajectory $i$ and search slot $k \in \{1,2,3\}$, collect the cumulative set of consumed retrieved passages $D_{i,k}$, deduplicated and sorted canonically by document ID, and select the shortest-token reference answer alias $y^*$.
\item \textbf{Cached potential evaluation}: Teacher-force $y^*$ under the frozen current policy $\pi_{\theta_0}$ conditioned on $C(x, D_{i,k})$ (Eq.~\ref{eq:potential}), reusing cached log-likelihoods whenever two prefixes share the same document-ID set within the batch.
\item \textbf{Return construction}: Compute slot-level returns over the $H=3$ horizon using either step-clipped increments $U^{\mathrm S}_{i,k}$ (Eq.~\ref{eq:v1}, ECA-Step) or horizon potential differences $U^{\mathrm H}_{i,k}$ (Eq.~\ref{eq:v2}, ECA-Horizon), padding unused trailing slots with unchanged evidence states ($\Delta_{i,j}=0$).
\item \textbf{Cross-fold normalization and novelty gating}: Partition the 8 rollouts per question into two random folds of 4, normalize each trajectory's return against the opposite fold's mean and floored standard deviation (Eq.~\ref{eq:norm}), and clamp positive auxiliary credit to zero if $D_{i,k}=D_{i,k-1}$ (Eq.~\ref{eq:gate}).
\item \textbf{Policy gradient step}: Add the scaled auxiliary credit $\lambda \widetilde Z_{i,k}$ ($\lambda=0.1$) to all policy tokens in the $k$-th consumed search action (both \texttt{<think>} and \texttt{<search>} tokens), accumulate the surrogate loss (Eq.~\ref{eq:loss}) across microbatches, update $\theta$, and refresh $\pi_{\theta_0} \leftarrow \pi_\theta$ for the next update.
\end{enumerate}

\subsection{Benchmark Sizes}
\label{app:scoring}
The seven Search QA test sets contain 51,713 questions in total: NQ \citep{kwiatkowski-etal-2019-natural} (3,610), TriviaQA \citep{joshi-etal-2017-triviaqa} (11,313), PopQA \citep{mallen-etal-2023-trust} (14,267), HotpotQA \citep{yang-etal-2018-hotpotqa} (7,405), 2WikiMultiHopQA \citep{ho-etal-2020-constructing} (12,576), MuSiQue \citep{trivedi-etal-2022-musique} (2,417), and Bamboogle \citep{press-etal-2023-measuring} (125).

\subsection{MultiHop-RAG Corpus and Transfer Protocol}
For MultiHop-RAG transfer, we segment all 609 source news articles into 256-token spans (under the E5 tokenizer) with a 32-token overlap while preserving article titles and metadata headers, producing 6,532 index chunks. All models are evaluated zero-shot with top-3 E5-base-v2 retrieval, greedy decoding, a maximum of 3 searches (4 actions), and a 16,384-token context limit.

\input{tables/related_mechanisms}

\section{Formal Properties of the Auxiliary Signal}
\label{app:properties}
\paragraph{Invariance to reasoning and query rephrasing.}
For any fixed question $x$, canonical target answer $y^*$, frozen scorer $\pi_{\theta_0}$, and canonicalized document set $D$, the potential $\Phi_i(D)$ in Eq.~\ref{eq:potential} depends exclusively on $(x, D, y^*)$. Consequently, altering the generated chain-of-thought or search query syntax without changing the retrieved document IDs leaves $\Phi_i(D)$ and $\Delta_{i,k}$ strictly unchanged.

\paragraph{Equivalence of ECA-Step and ECA-Horizon without saturation.}
Whenever $|\Delta_{i,j}| \le c$ for all $j \ge k$, the step-wise clip in Eq.~\ref{eq:v1} acts as the identity, reducing $U^{\mathrm S}_{i,k}$ to $(cH)^{-1}\sum_{j=k}^H \Delta_{i,j} = U^{\mathrm H}_{i,k}$ (Eq.~\ref{eq:v2}). If $|\Delta_{i,j}| \le c$ holds across all trajectories and slots in a question group, ECA-Step and ECA-Horizon produce identical normalized advantages $\widetilde Z_{i,k}$. Notably, even if a target trajectory's own increments lie within $[-c,c]$, saturation in an opposite-fold trajectory can still change the reference mean $\mu_{-f(i),k}$ or standard deviation $s_{-f(i),k}$, causing the normalized advantages of the two variants to diverge.

\paragraph{Boundedness and signal recovery in all-failure groups.}
By construction (Eqs.~\ref{eq:norm}--\ref{eq:combined}), the auxiliary token correction satisfies $|\lambda \widetilde Z_{i,k}| \le 2\lambda = 0.2$ for every search token, and equals $0$ whenever all rollouts in a group collect identical evidence returns ($U_{i,k} = \mu_{-f(i),k}$). When all $G$ rollouts fail to match the terminal answer ($R_1 = \dots = R_G = 0$), $A_i^{\mathrm{out}} = 0$, so the policy gradient is driven purely by differences in $\widetilde Z_{i,k}$ across search actions. Because $\widetilde Z_{i,k}$ is nonlinearly normalized, clipped, and masked to search actions over trajectory-dependent states, it serves as a bounded process-shaping residual rather than an action-independent potential-based shaping term.

\paragraph{Why full-group normalization attenuates sparse evidence spikes.}
In terminal outcome GRPO (Eq.~\ref{eq:outcome}), rewards $R_i \in \{0,1\}$ are binary, so all successful rollouts share the same unit scale. Intermediate evidence returns $U_{i,k}$, by contrast, are continuous and often sparse on difficult multi-hop questions: suppose a single rollout discovers a high-value passage with return $u > 0$ while the remaining $G-1$ rollouts yield $0$, giving the group vector $(u, 0, \dots, 0) \in \mathbb{R}^G$. Under full-group normalization across all $G$ trajectories (including the evaluated trajectory itself), the sample mean is $\bar U = u/G$ and the unbiased sample standard deviation is $s_U = u/\sqrt{G}$. Whenever $u/\sqrt{G} \ge \sigma_{\min}$, the self-included z-score of the successful search becomes
\begin{equation}
\frac{u - \bar U}{s_U} = \frac{u - u/G}{u/\sqrt{G}} = \left(1 - \frac{1}{G}\right)\sqrt{G},
\end{equation}
in which the return magnitude $u$ cancels out completely between the numerator and denominator, assigning the exact same pre-clip advantage whether the retrieved passage yields a modest likelihood gain or a decisive breakthrough. Normalizing each trajectory against the disjoint opposite fold ($\mu_{-f(i),k}, s_{-f(i),k}$ in Eq.~\ref{eq:norm}) decouples the baseline statistics from $U_{i,k}$ itself. Empirically, replacing $4+4$ cross-fold normalization in 3B ECA-Step with full-group normalization across all $G=8$ trajectories (keeping zero-padding and novelty gating identical) reduces test macro EM by $2.04$ pp from $38.05$ to $36.01$.

%% file: sections/behavior.tex
\section{Behavior Before and After Training}
\label{app:behavior}
\input{tables/behavior_before_after}
\input{tables/behavior_paired}
To examine how reinforcement learning changes agent interaction patterns beyond exact-match accuracy, we compare each selected checkpoint against the untrained Qwen2.5-Instruct Base model of the same size using the identical evaluation harness across all 51,713 questions. Table~\ref{tab:behavior} reports dataset-macro averages of the number of model actions (including search calls, invalid actions, and the terminal answer), executed retrieval calls, generated policy tokens (excluding prompts and retrieved passages), and the fraction of episodes containing at least one malformed action.

\paragraph{Interaction budget, query repetition, and output length.}
Before RL training, the Base agents average $3.36$ actions ($2.28$ searches) at 3B and $2.66$ actions ($1.64$ searches) at 7B, with malformed action rates of $10.62\%$ and $1.44\%$, respectively. After RL training, GRPO, RLOO, ECA-Step, and ECA-Horizon all converge to the full interaction budget of $4.00$ actions and $3.00$ searches per question while driving invalid action rates close to zero ($\le 0.44\%$). At 3B, however, expanding from $2.28$ to $3.00$ searches under outcome-only supervision causes the policy to frequently re-issue identical queries: the episode-level duplicate-query rate rises from $23.86\%$ in the Base agent to $42.40\%$ under GRPO and $54.07\%$ under RLOO. By pairing step-local evidence credit with novelty gating ($D_{i,k}=D_{i,k-1} \implies \widetilde Z_{i,k} \le 0$), ECA suppresses 3B query repetition to $30.26\%$ (ECA-Step) and $24.84\%$ (ECA-Horizon), matching the Base model's repetition rate despite executing $3.00$ searches per episode (at 7B, the stronger backbone already maintains low query repetition below $8.5\%$ across all methods). Output token counts also show that verbosity alone does not dictate accuracy: at 3B, ECA-Horizon achieves higher macro EM than ECA-Step ($38.22$ vs.\ $38.05$) while generating $22\%$ fewer policy tokens per episode ($200.9$ vs.\ $257.1$), whereas at 7B ECA-Horizon generates slightly more tokens than ECA-Step ($92.0$ vs.\ $83.6$) and fewer than RLOO ($158.6$).

\paragraph{Paired question-level transitions.}
Table~\ref{tab:paired-behavior} tracks question-by-question correctness transitions from the untrained Base model to ECA across the pooled 51,713 test examples. At 3B, ECA-Horizon corrects $12,340$ questions ($23.86\%$) that the Base agent answered incorrectly, while regressing on $2,166$ previously correct questions ($4.19\%$), yielding a net pooled gain of $+19.67$ pp. At 7B, ECA-Horizon converts $8,861$ wrong answers to correct ($17.13\%$) against $2,158$ regressions ($4.17\%$), for a net pooled gain of $+12.96$ pp.

%% file: tables/behavior_before_after.tex
\begin{table*}[t]
\caption{Agent interaction statistics before and after RL across the seven Search QA benchmarks (unweighted macro averages over 51,713 questions). $\Delta$ EM is measured relative to the same-size Base agent. Actions count all model turns (including invalid actions and the final answer); Searches count executed retrieval calls; Dup.\ Query is the percentage of episodes containing at least one repeated search query; Tokens count generated policy tokens (excluding prompts and retrieved text); and Invalid is the percentage of episodes with at least one malformed action.}
\label{tab:behavior}
\centering\small
\setlength{\tabcolsep}{7pt}
\begin{tabular}{lrrrrrrr}
\toprule Method & EM (\%) & $\Delta$ EM (pp) & Actions & Searches & Dup.\ Query (\%) & Tokens & Invalid (\%) \\
\midrule
\multicolumn{8}{l}{\textbf{Qwen2.5-3B-Instruct}} \\
Base agent & 21.75 & +0.00 & 3.36 & 2.28 & 23.86 & 104.2 & 10.62 \\
GRPO & 37.19 & +15.44 & 4.00 & 3.00 & 42.40 & 178.9 & 0.00 \\
RLOO & 35.43 & +13.68 & 4.00 & 3.00 & 54.07 & 120.8 & 0.02 \\
ECA-Step & \underline{38.05} & +16.29 & 4.00 & 3.00 & 30.26 & 257.1 & 0.07 \\
ECA-Horizon & \textbf{38.22} & +16.46 & 4.00 & 3.00 & 24.84 & 200.9 & 0.14 \\
\midrule
\multicolumn{8}{l}{\textbf{Qwen2.5-7B-Instruct}} \\
Base agent & 29.93 & +0.00 & 2.66 & 1.64 & 1.69 & 72.1 & 1.44 \\
GRPO & 37.82 & +7.89 & 4.00 & 3.00 & 0.71 & 76.7 & 0.06 \\
RLOO & 40.56 & +10.63 & 4.00 & 3.00 & 7.07 & 158.6 & 0.02 \\
ECA-Step & \underline{41.65} & +11.72 & 4.00 & 3.00 & 8.37 & 83.6 & 0.44 \\
ECA-Horizon & \textbf{42.62} & +12.69 & 4.00 & 3.00 & 8.42 & 92.0 & 0.01 \\
\bottomrule
\end{tabular}
\end{table*}

%% file: tables/behavior_paired.tex
\begin{table*}[t]
\caption{Paired question-level correctness transitions from the untrained Base agent to ECA across all 51,713 pooled test questions (count and percentage of total questions). $\Delta$ acc.\ denotes the net change in pooled accuracy, which weights all questions equally rather than weighting the seven datasets equally.}
\label{tab:paired-behavior}
\centering\small
\setlength{\tabcolsep}{4pt}
\begin{tabular}{llrrrrr}
\toprule Model & Variant & Wrong $\to$ correct & Correct $\to$ wrong & Both correct & Both wrong & $\Delta$ acc. (pp) \\
\midrule
3B & ECA-Step & 12,115 (23.43) & 2,081 (4.02) & 10,186 (19.70) & 27,331 (52.85) & +19.40 \\
3B & ECA-Horizon & 12,340 (23.86) & 2,166 (4.19) & 10,101 (19.53) & 27,106 (52.42) & +19.67 \\
7B & ECA-Step & 8,473 (16.38) & 2,320 (4.49) & 15,275 (29.54) & 25,645 (49.59) & +11.90 \\
7B & ECA-Horizon & 8,861 (17.13) & 2,158 (4.17) & 15,437 (29.85) & 25,257 (48.84) & +12.96 \\
\bottomrule\end{tabular}
\end{table*}

%% file: tables/related_mechanisms.tex
\begin{table*}[t]
\caption{Mechanism-level comparison of intermediate credit assignment and group optimization methods for search agents.}
\label{tab:related-mechanisms}
\centering\small
\setlength{\tabcolsep}{4pt}
\begin{tabular}{>{\raggedright\arraybackslash}p{0.18\textwidth}>{\raggedright\arraybackslash}p{0.27\textwidth}>{\raggedright\arraybackslash}p{0.23\textwidth}>{\raggedright\arraybackslash}p{0.23\textwidth}}
\toprule
Method & Intermediate signal & Additional source or computation & Distinction from ECA \\
\midrule
GiGPO \citep{feng2025gigpo} & Relative returns at repeated/similar states & State grouping across sampled trajectories & Groups outcomes, rather than scoring answer likelihood from document sets. \\
IGPO \citep{wang2026igpo} & Mean answer log-likelihood change on rollout prefixes & Current-policy teacher forcing; separate process/outcome normalization & Normalizes immediate rewards before discounted accumulation; ECA constructs returns before cross-fold normalization. \\
TIPS \citep{xie2026tips} & Answer-likelihood potential differences & Frozen/refreshed teacher on dialogue context & Uses dialogue-context potentials; ECA restricts its readout to questions and documents and applies a search-local mask. \\
StepSearch \citep{zheng-etal-2025-stepsearch} & Gold-document coverage and repetition penalties & Sub-question/search supervision and evidence references & ECA uses answer labels without gold supporting-passage annotations. \\
PiCA \citep{liu2026pica} & History-dependent success-potential progress & Pivot-annotated trajectories; trained reward model & ECA reuses the current actor for a document-only readout. \\
KnowGPO \citep{chen-etal-2026-knowgpo} & Gains in satisfied atomic knowledge conditions & Constructed conditions and LLM verification & ECA requires no external LLM verifier or constructed atomic conditions, relying only on document-ID deduplication and current-policy answer scoring. \\
Retrieval credit \citep{huang2026retrievalcredit} & Coverage, dependency-aware coverage, answer matching & Corpus-grounded evidence annotations & Compares scalar versus query-aligned incorporation; ECA's residual covers the full search action. \\
CW-GRPO \citep{wang2026cwgrpo} & LLM-judged round contributions & Retrieval/reasoning judge & Reweights outcome advantages; ECA adds a bounded residual. \\
IGSD \citep{jiang2026igsd} & Verified utility difference between paired queries & Successful-sibling hindsight and counterfactual query execution & We score consumed evidence without donor-conditioned distillation or new query continuations. \\
\bottomrule
\end{tabular}
\end{table*}

%% file: sections/related_comparison.tex
\section{Comparison with Related Credit Mechanisms}
\label{app:related-comparison}
Table~\ref{tab:related-mechanisms} contrasts ECA with closely related multi-turn credit assignment and group optimization methods along three axes: the source of intermediate supervision, its computational or annotation requirements relative to outcome-only rollouts, and how step-level signals enter the policy update. Whereas corpus-supervised methods rely on gold supporting-document labels or learned reward models, and dialogue-prefix likelihood methods (IGPO, TIPS) evaluate answer probability on the full interaction history, ECA combines an evidence-only likelihood readout with cross-fold return normalization, novelty gating, and search-action-local assignment.

%% file: sections/case_studies.tex
\section{Qualitative Comparisons with GRPO}
\label{app:cases}
\input{tables/case_iso_bridge}
\input{tables/case_award_year}
\input{tables/case_lifespans}
Tables~\ref{tab:case-iso}--\ref{tab:case-lifespans} present three representative test cases in which ECA-Horizon answers correctly while outcome-only GRPO fails, illustrating multi-hop bridge-entity retrieval, temporal disambiguation, and decomposed entity searches.

Each comparison uses the same model size, retrieval corpus, and evaluation harness with the development-selected checkpoints in Appendix~\ref{app:protocol}. All six trajectories execute three searches with no invalid actions. We report the issued queries, relevant excerpts from retrieved passages, final answers, and normalized EM, where \emph{New passage IDs per search} counts retrieved document IDs not seen in earlier turns of the trajectory.

%% file: tables/case_iso_bridge.tex
\begin{table*}[t]
\caption{\textbf{Following a bridge entity (3B, MuSiQue).} Both models retrieve the country name. GRPO repeats the initial query and returns this intermediate entity; ECA-Horizon follows it to a passage stating the country's location.}
\label{tab:case-iso}
\centering
\small
\setlength{\tabcolsep}{5pt}
\renewcommand{\arraystretch}{1.14}
\begin{tabular}{@{}>{\raggedright\arraybackslash}p{0.48\textwidth}>{\raggedright\arraybackslash}p{0.48\textwidth}@{}}
\toprule
\multicolumn{2}{@{}p{0.975\textwidth}@{}}{\textbf{Question:} Where is the country with ISO code ISO 3166-2:CV located?\newline
\textbf{Gold:} central Atlantic Ocean. \quad \textbf{ID:} \texttt{dev\_88}.} \\
\midrule
\textbf{GRPO} & \textbf{ECA-Horizon} \\
\midrule
\textbf{1.} \texttt{Where is the country with ISO code ISO 3166-2:CV located}\newline
\textbf{2.} \texttt{Where is the country with ISO code ISO 3166-2:CV located}\newline
\textbf{3.} \texttt{Where is the country with ISO code ISO 3166-2:CV located} & \textbf{1.} \texttt{Where is the country with ISO code ISO 3166-2 CV located}\newline
\textbf{2.} \texttt{Cabo Verde location}\newline
\textbf{3.} \texttt{Cabo Verde location map} \\
\midrule
\textbf{Search 1:} \emph{ISO 3166-2:CV} (passage 18933183).\newline``ISO 3166-2:CV is the entry for Cabo Verde in ISO 3166-2'' & \textbf{Search 1:} \emph{ISO 3166-2:CV} (passage 18933183).\newline``ISO 3166-2:CV is the entry for Cabo Verde in ISO 3166-2''\par\smallskip
\textbf{Search 2:} \emph{Cape Verde} (passage 5648183).\newline``an island country spanning an archipelago of 10 volcanic islands in the central Atlantic Ocean.'' \\
\midrule
\textbf{New passage IDs per search:} 3 / 0 / 0 & \textbf{New passage IDs per search:} 3 / 3 / 1 \\
\textbf{Final answer:} Cabo Verde \quad ($\mathrm{EM}=0$) & \textbf{Final answer:} central Atlantic Ocean \quad ($\mathrm{EM}=1$) \\
\bottomrule
\end{tabular}
\end{table*}

%% file: tables/case_award_year.tex
\begin{table*}[t]
\caption{\textbf{Resolving the award year before the club year (3B, Bamboogle).} GRPO follows a retrieved passage about the 2012 award to Messi. ECA-Horizon retrieves the 2007 winner, Kaká, and a passage linking him to Real Madrid in 2011--12.}
\label{tab:case-award}
\centering
\small
\setlength{\tabcolsep}{5pt}
\renewcommand{\arraystretch}{1.14}
\begin{tabular}{@{}>{\raggedright\arraybackslash}p{0.48\textwidth}>{\raggedright\arraybackslash}p{0.48\textwidth}@{}}
\toprule
\multicolumn{2}{@{}p{0.975\textwidth}@{}}{\textbf{Question:} For which club did the winner of the 2007 Ballon d'Or play for in 2012?\newline
\textbf{Gold:} Real Madrid. \quad \textbf{ID:} \texttt{test\_104}.} \\
\midrule
\textbf{GRPO} & \textbf{ECA-Horizon} \\
\midrule
\textbf{1.} \texttt{2007 Ballon d'Or winner club 2012}\newline
\textbf{2.} \texttt{Lionel Messi club 2012}\newline
\textbf{3.} \texttt{Lionel Messi club 2012} & \textbf{1.} \texttt{For which club did the winner of the 2007 Ballon d'Or play for in 2012}\newline
\textbf{2.} \texttt{2007 Ballon d'Or Kaká club 2012}\newline
\textbf{3.} \texttt{Kaká 2012 club} \\
\midrule
\textbf{Search 1:} \emph{2012 FIFA Ballon d'Or} (passage 17186262).\newline``The odds-on favourite Lionel Messi won the award.'' & \textbf{Search 1:} \emph{2007 Ballon d'Or} (passage 20994564).\newline``was awarded to Kaká.''\par\smallskip
\textbf{Search 3:} \emph{Kaká} (passage 13277486).\newline``In 2011--12, Real Madrid won La Liga with a record 100 points that year, with Kaká providing nine assists'' \\
\midrule
\textbf{New passage IDs per search:} 3 / 3 / 0 & \textbf{New passage IDs per search:} 3 / 3 / 2 \\
\textbf{Final answer:} Barcelona \quad ($\mathrm{EM}=0$) & \textbf{Final answer:} Real Madrid \quad ($\mathrm{EM}=1$) \\
\bottomrule
\end{tabular}
\end{table*}

%% file: tables/case_lifespans.tex
\begin{table*}[t]
\caption{\textbf{Separating two entity searches (7B, 2WikiMultiHopQA).} GRPO's repeated joint query retrieves an unrelated longevity record. ECA-Horizon retrieves both biographies; their returned dates support the correct lifespan comparison.}
\label{tab:case-lifespans}
\centering
\small
\setlength{\tabcolsep}{5pt}
\renewcommand{\arraystretch}{1.14}
\begin{tabular}{@{}>{\raggedright\arraybackslash}p{0.48\textwidth}>{\raggedright\arraybackslash}p{0.48\textwidth}@{}}
\toprule
\multicolumn{2}{@{}p{0.975\textwidth}@{}}{\textbf{Question:} Who lived longer, Horace Lawson Hunley or Ernst Ziehm?\newline
\textbf{Gold:} Ernst Ziehm. \quad \textbf{ID:} \texttt{dev\_12551}.} \\
\midrule
\textbf{GRPO} & \textbf{ECA-Horizon} \\
\midrule
\textbf{1.} \texttt{Horace Lawson Hunley lifespan Ernst Ziehm lifespan}\newline
\textbf{2.} \texttt{Horace Lawson Hunley lifespan Ernst Ziehm lifespan}\newline
\textbf{3.} \texttt{Horace Lawson Hunley death Ernst Ziehm death} & \textbf{1.} \texttt{Horace Lawson Hunley}\newline
\textbf{2.} \texttt{Ernst Ziehm}\newline
\textbf{3.} \texttt{Horace Lawson Hunley died} \\
\midrule
\textbf{Search 1:} \emph{Waldo McBurney} (passage 13943908).\newline``Ralph Waldo McBurney (October 3, 1902 -- July 8, 2009)'' & \textbf{Search 1:} \emph{Horace Lawson Hunley} (passage 14541340).\newline``June 20, 1823 in Sumner County, Tennessee -- October 15, 1863 off Charleston, South Carolina''\par\smallskip
\textbf{Search 2:} \emph{Ernst Ziehm} (passage 7050528).\newline``1 May 1867 -- 7 July 1962'' \\
\midrule
\textbf{New passage IDs per search:} 3 / 0 / 3 & \textbf{New passage IDs per search:} 3 / 3 / 1 \\
\textbf{Final answer:} Horace Lawson Hunley \quad ($\mathrm{EM}=0$) & \textbf{Final answer:} Ernst Ziehm \quad ($\mathrm{EM}=1$) \\
\bottomrule
\end{tabular}
\end{table*}